\documentclass[11pt]{article}
\usepackage{acl}
\usepackage{microtype}
\usepackage{graphicx} 
\usepackage{float}
\usepackage{booktabs}
\usepackage{appendix}
\usepackage{longtable}
\usepackage{placeins}
\usepackage{amsmath}
\usepackage{multirow}
\usepackage{enumitem}
 
\usepackage[dvipsnames]{xcolor}

\usepackage{siunitx}
\title{Heads, Tails, and AI Fails: LLMs, Randomness, and Human Judgments}

\author{
	Katherine Van Koevering \\
	Data Science and AI Institute \\
	Johns Hopkins University \\
	Baltimore, MD, USA
	\And
	Jon Kleinberg \\
	Departments of Computer Science \\
	and Information Science \\
	Cornell University \\
	Ithaca, NY, USA
}
\begin{document}
\maketitle
\begin{abstract}
Randomness is central to human cognition and to many applications in which large language models are deployed, yet probabilistic token generation does not imply that LLMs can produce unbiased random sequences. We study how contemporary LLMs generate binary random sequences using the classic behavioral-science paradigm of simulated coin flips. Across single flips, 20-flip sequences, n-gram statistics, run lengths, alternation rates, and next-flip predictability, we compare model outputs to both true Bernoulli baselines and human data from prior work. We find that LLMs reproduce several canonical human randomness biases, including over-alternation, aversion to long runs, and first-flip biases, but often amplify them or introduce model-specific distortions. Increasing temperature reduces some rigid patterns but does not eliminate systematic structure. We further investigate over-alternation through prompt-framing experiments, continuation tests, corpus searches, and residual-stream probes, finding evidence that this bias is not explained by direct memorization or tokenization alone. These results show that LLMs are unreliable generators of randomness and imperfect simulators of human randomness judgments, with implications for applications requiring stochasticity, fair sampling, or behavioral simulation. 
\end{abstract}

\section{Introduction}

Understanding LLM performance on predictions of randomness offers a valuable test setting for LLM mimicry of humans as well as informs a variety of LLM applications. Few things are as consistently human as our ability to perceive patterns. Inversely, humans are generally terrible at recognizing (and producing) randomness. Our failures with randomness are predictable, consistent, and measurable in a way that few human biases are. These systematic failures of randomness are one of the most well studied and consistent phenomena of human cognition, with well over a hundred years of research stretching from philosophy to physics \cite{Nickerson2002, ford1983random, ayer2006probability, Goodfellow1940, Dennett1991, Alberoni1962,falk1991, Kac1983}. Dozens of studies have failed to find any factor that ameliorates our failures with randomness. The combination of consistency and extensive documentation makes randomness production an ideal setting in which to analyze LLM mimicry of human biases. Given that both a perfect non-human response (uniform randomness) and a perfect human response (as studied for over a century by behavioral scientists) are extraordinarily well-defined, this problem allows detailed analysis of the extent to which various production models reproduce these fallacies. The most common set-up for human experiments is the production of “random” binary sequences, which we mimic here for easy comparison to the extensive historical analysis of human behavior. 

\begin{figure*}[h!]
    \centering
    \includegraphics[width = \textwidth]{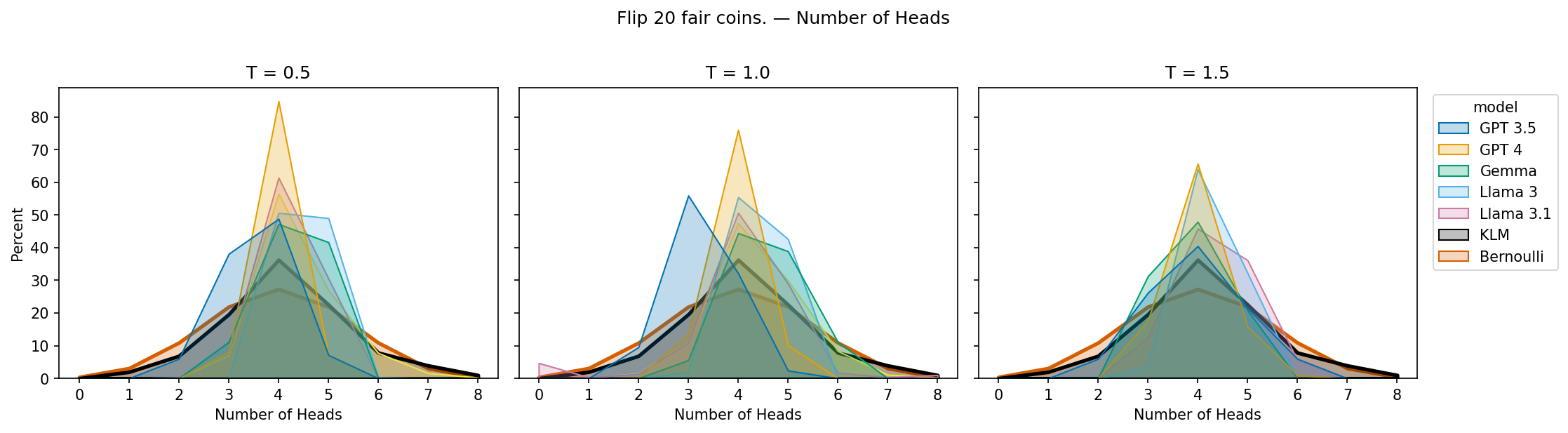}
    \caption{Percentage of runs with $x$ heads for different models, using the 'Flip 20 fair coins' prompt. The data for KLM is taken from Kleinberg et al. \cite{kleinberg2017theory} that performed similar experiments looking at sequences of 8 coin flips taken from longer samples. Bernoulli represents a `truly random' sequence of Bernoulli distributions. Note that KLM is most similar to the Bernoulli distribution, but that all models, even at higher temperatures, exhibit and exacerbate the human pattern of too many flips with nearly half heads and half tails and too few at the extremes.}
    \label{fig:num_heads}
\end{figure*}

Furthermore, numerous proposed LLM use cases explicitly or implicitly assume  models are capable of producing randomness: agents are asked to simulate human behavior in psychology experiments~\cite{ke2024exploring}, to play games that rely on stochasticity such as rock-paper-scissors~\cite{wu2023smartplay}, or to act as assistants in creative or exploratory environments such as Dungeons and Dragons~\cite{zhu2023calypso} or Minecraft~\cite{wang2023voyager}. In reinforcement learning and bandit problems, randomness is fundamental to exploration strategies~\cite{krishnamurthy2024can, nie2024evolve}. Decision-based models such as DeLLMa~\cite{liu2024dellma} may optimize utility but still require randomization to select among options.

In this work, we explore how LLMs approach randomness and where and how they fail. There are three possibilities: (1) LLMs could behave like humans, replicating human biases in randomness generation; (2) they could be further than humans from uniform randomness, producing more deterministic or structured outputs; or (3) they could be closer than humans to uniform randomness, introducing noise beyond human tendencies. Each scenario has implications. If LLMs replicate human biases, they may serve as useful simulators of human cognition, but also inherit human failures. If they are more random, they may fail to capture human-like heuristics, limiting their use in simulation studies, but be more useful in tasks requiring pure randomness or games. If they are less random, they may be unsuitable for tasks requiring stochasticity. Additionally, this dichotomy of humanness has previously been explored as one aspect of measuring advanced AI~\cite{PERSAUD2005211}.



This paper makes four contributions. First, we provide a systematic evaluation of LLM-generated binary random sequences across model families, prompts, and decoding temperatures. Second, we compare model outputs to both IID Bernoulli baselines and human behavioral data, showing that LLMs reproduce but often amplify canonical human randomness biases, including over-alternation, aversion to long runs, over-balancing, and first-flip effects. Third, we show that these biases remain even when temperature is increased, implying that stochastic decoding alone is not a reliable route to unbiased randomization. Fourth, we investigate possible sources of the dominant over-alternation bias through prompt-framing experiments, continuation tests, corpus searches, and internal probing, finding evidence against simple explanations based solely on prompt wording, direct memorization, or tokenization. 

\section{Related Work}

\paragraph{Human Randomness Generation}
Humans systematically fail to produce random sequences, a finding replicated across over a century of research \cite{Nickerson2002, ford1983random, Goodfellow1940, Tune1964}. Human-generated sequences differ markedly from IID processes \cite{Budescu, Falk1997, Allen1986, treisman1987}: people over-alternate between outcomes (the \textit{negative recency effect}) \cite{budescu1985analysis, falk1981perception, Rapoport1992, Bar-Hillel1991}, avoid long runs \cite{Goodfellow1940, Bakan1960}, over-balance symbol frequencies \cite{Nickerson2009}, and exhibit first-flip biases \cite{Bar-Hillel2014}. These patterns persist across variations in instructions \cite{Bar-Hillel2014, Rapoport1992}, education \cite{Bakan1960}, stress \cite{Heuer2005}, and clinical populations \cite{Weiss1965}, though some biases strengthen under cognitive load \cite{Heuer2005}. For comprehensive reviews, see \citet{Nickerson2002} and \citet{Tune1964}.

\paragraph{LLMs and Randomness}
Recent work reveals that LLMs also fail to produce unbiased random outputs. Models favor particular digits and avoid round numbers in random-number tasks \cite{hopkins2023can, sedov2025polariton}, and deviate from uniform mixed strategies in games like Rock-Paper-Scissors despite correctly stating the theoretical solution \cite{guo2025illusion, vidler2025playing}. Temperature tuning and prompt redesign shift but do not eliminate these biases \cite{vidler2025playing}. Entropy metrics and NIST randomness tests confirm that LLM outputs frequently fail standard randomness criteria \cite{karanjai2025evaluating}. Our work extends this literature by directly comparing LLM-generated sequences to both Bernoulli baselines and human behavioral data, enabling analysis of whether LLM biases mirror, amplify, or diverge from known human heuristics.

\section{LLMs, Randomness, and Human Judgments}

Unlike humans, when LLMs are used for almost any task, the temperature variable is a key parameter. Higher temperature seems likely to give more random (and thus less consistent) results. However, in tests, setting temperature above 1.5 proved impractical, as the model began to return results that held random text and we were unable to parse a coin flip from the text. As such, we test for temperatures from 0 to 1.5.\footnote{We treat temperature 0 as a practical low-stochasticity setting rather than a formal guarantee of determinism, since proprietary APIs may introduce backend nondeterminism, load-balancing effects, or undocumented decoding behavior. }. 

Although LLMs define probability distributions over next tokens, their outputs are not random in the sense required by a Bernoulli coin-flip process. Given a prompt, the model produces logits over the vocabulary. Decoding then converts these logits into a sampled token, with temperature rescaling the distribution before sampling. At low temperature, decoding concentrates probability mass on the highest-probability tokens and may become effectively deterministic; at high temperature, lower-probability tokens become more likely, but the distribution remains shaped by the model’s learned logits and by additional decoding parameters such as top-p or top-k sampling. Thus, unbiased random generation requires more than sampling noise: the model must assign appropriate probability mass to the relevant outcomes, and the decoding procedure must preserve that balance.  

\subsection{Single Coin Flips}
We prompted several LLMs with the instructions ``Flip a coin.'' and ``Flip a fair coin.'' \footnote{Gemma refused both of these prompts claiming not to be capable of random number generation. We included the word "simulate" to bypass this}, sampling outputs across temperatures from 0.0 to 1.5, with at least 100 repetitions per model.

\begin{table*}[]
    \centering
    \begin{tabular}{lrrrrrlrr}
t & Gemma & GPT 3.5 & GPT 4 & Lllama 3 & Llama 3.1  &Qwen& Human & Bernoulli\\
0.0 & 1.00 & 1.00 & 0.65 & 1.00 & 1.00  &0.81& 0.80 & 0.50 \\
0.5 & 1.00 & 1.00 & 0.93 & 1.00 & 1.00  &0.80& 0.80 & 0.50 \\
1.0 & 1.00 & 0.90 & 0.89 & 1.00 & 1.00  &0.81& 0.80 & 0.50 \\
1.5 & 0.79 & 0.86 & 0.80 & 0.97 & 1.00  &0.74& 0.80 & 0.50 \\
\end{tabular}
    \caption{Proportion of responses that are heads for different models and temperatures, using the prompt ``flip a fair coin''}
    \label{tab:single_flip}
\end{table*}

\textbf{LLM vs True Randomness.} A truly random process would yield approximately 50\% heads and 50\% tails. However, Llama-3 exhibited a striking deviation: across 660 flips from 0.0 to 1.0, it produced 100\% heads. Only at temperature 1.5 did tails appear, and even then, at extremely low rates. This pattern persisted in Gemma and Llama 3.1, which also failed to produce any tails across all tested conditions. Qwen, however, was lightly lower - near human baseline.

\textbf{LLM vs Human Judgments.} Humans also exhibit a heads-first bias when simulating coin flips, with over 80\% of sequences beginning with heads~\cite{Bar-Hillel2014}. LLMs replicate this bias but exaggerate it. While humans still produce tails with substantial frequency, Llama-3 and Gemma effectively eliminate them. ChatGPT models behave differently: they show more variability and a higher rate of refusals, with some configurations producing more balanced outputs.

\subsection{Coin-flip Sequences}
Given the long history of binary random sequences, we borrow heavily from prior work on the subject. As such, for our multi-flip setup we ask LLMs to flip coins to generate the sequence. We use two prompts ``Flip 20 coins.'' and ``Flip 20 fair coins.'' Although we do see some difference between the prompts, for the most part the prompts give us remarkably similar results. For most analysis, we break these 20-flip sequences into 19 overlapping 8-flip sequences, similarly to Kleinberg et al \cite{kleinberg2017theory}. This allows direct comparison with several known papers who also report results for 8-flip sequences \cite{kleinberg2017theory, Lopes1987}. This also means that all subsequences of a given length are equally probable.
\begin{table*}[]
    \centering
\begin{tabular}{lrrrrrlrr}
t & Gemma & GPT 3.5 & GPT 4 & Lllama 3 & Llama 3.1  & Qwen & Human & Bernoulli \\
0.00 & 1.00 & 1.00 & 0.98 & 1.00 & 1.00  &0.96& 0.80 & 0.50 \\

0.5 & 1.00 & 1.00 & 0.89 & 1.00 & 1.00  &0.98& 0.80 & 0.50 \\

1.0 & 1.00 & 1.00 & 0.89 & 1.00 & 0.81  &0.94& 0.80 & 0.50 \\
1.5 & 0.94 & 0.91 & 0.89 & 1.00 & 0.77  &-& 0.80 & 0.50 \\
\end{tabular}
    \caption{Proportion of first responses in a sequence that are heads for different models and temperatures, using the prompt ``flip 20 fair coins''.}
    \label{tab:first_flip}
\end{table*}
\subsubsection{First Flip Bias}

\textbf{LLM vs True Randomness.}Aggregated across all valid model-temperature cells, over 94\% of sequences began with heads  (Tab.~\ref{tab:first_flip}).  Llama 3.1 and GPT-4 were somewhat less biased than others, but the overall skew remained. These results deviate significantly from the 50\% expectation of a fair coin.

\textbf{LLM vs Human Judgments.} Humans also show a heads-first bias, but to a lesser extent~\cite{Bar-Hillel2014}. Prompt ordering affects human responses: 87\% begin with heads when “heads” is mentioned first, but only 67\% do so when “tails” is listed first. LLMs do not show the same human-like primacy effect. Instead, outcome-order manipulations produce heterogeneous and sometimes reversed effects across models. Qwen and Llama 3, for example, produce more heads when tails is mentioned first, suggesting model-specific recency or prompt-format effects rather than a simple human-like primacy bias. Qwen is particularly sensitive to outcome ordering, with the probability that the generated sequence begins with H rising from 0.474 under H-first wording to 0.912 under T-first wording. 

\textbf{Synthesis.} LLMs replicate the human heads-first bias but lack the prompt sensitivity that characterizes human responses. Their behavior appears driven more by linguistic artifacts (e.g., frozen binomials~\cite{van2020frozen}) than by human-like heuristics or stochastic processes.

\begin{table}[ht]
\centering
\small
\caption{Proportion of sequences beginning with \texttt{H} under heads-first vs.\
 tails-first outcome ordering, \textit{Flip 20 coins.}\ at $T{=}1.5$.}
\label{tab:ordering}
\begin{tabular}{lccc}
\toprule
\textbf{Model} & \textbf{H-first} & \textbf{T-first} & $\boldsymbol{\Delta}$ \\
\midrule
GPT-4           & 0.846 & 0.897 & $+0.051$ \\
GPT-3.5         & 0.762 & 0.500 & $-0.262$ \\
Llama~3 (70B)   & 0.500 & 0.800 & $+0.300$ \\
Llama~3.1 (70B) & 0.611 & 0.579 & $-0.032$ \\
Gemma~2 (27B)   & 0.789 & 0.938 & $+0.148$ \\
Qwen~2 (7B)     & 0.474 & 0.912 & $+0.438$ \\
\bottomrule
\end{tabular}
\end{table}


\subsubsection{Fraction of Heads}

\textbf{LLM vs True Randomness.} In an 8-flip sequence, a fair coin yields a binomial distribution centered at 4 heads. LLMs deviate from this expectation by overproducing sequences with moderate head counts and underproducing extremes. As shown in Figure~\ref{fig:run_frac}, models like Gemma, GPT-4, Llama 3, and Llama 3.1 tend to cluster around 4 heads per sequence, especially at higher temperatures. At lower temperatures, Llama 3 and Llama 3.1 show a slight but significant bias toward heads, while GPT-3.5 exhibits a strong tails bias that diminishes as temperature increases.

\textbf{LLM vs Human Judgments.} Humans are known to overcorrect for randomness, producing sequences that are “too balanced” compared to true stochastic processes~\cite{Nickerson2009}. LLMs replicate and even amplify this tendency. At high temperatures, most models average just over 4 heads per 8 flips — closely mirroring human behavior, but with even tighter clustering. GPT-3.5 again diverges: at low temperatures, it shows a tails bias not typically observed in human data, though it converges toward human-like distributions at higher temperatures.

\subsubsection{Runs and N-Grams}

\textbf{LLM vs True Randomness.} In a fair coin sequence, we expect an average of 3.5 alternations in 8 flips, a predictable distribution of run lengths, and n-gram frequencies consistent with a Bernoulli process. LLMs deviate from all three.

\begin{figure*}
    \centering
    \includegraphics[width=\textwidth]{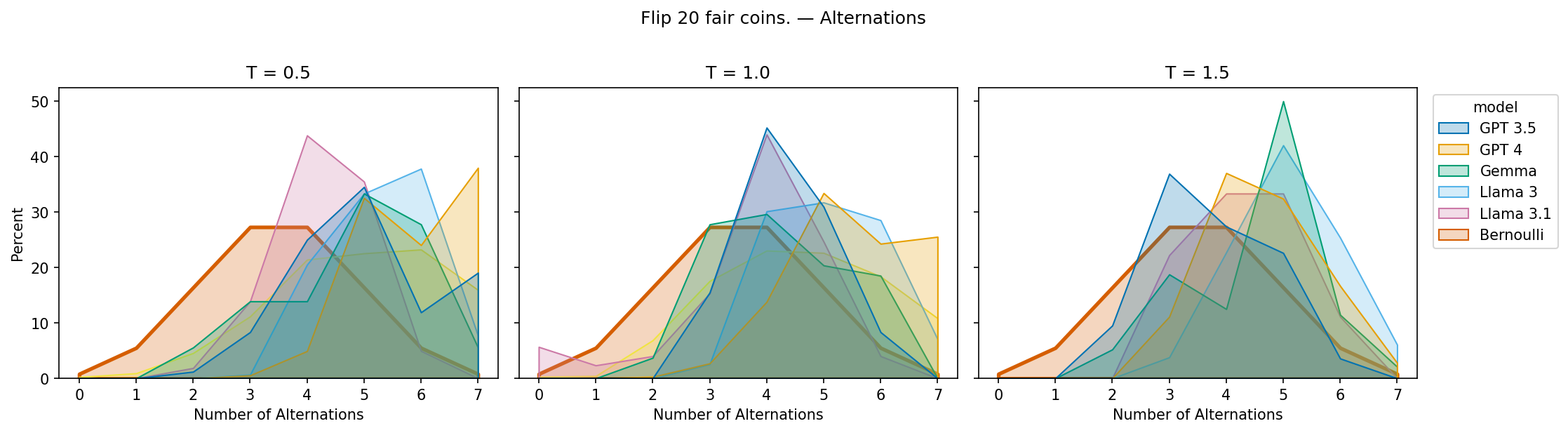}
    \caption{Histograms of the number of alternations in an 8-flip sequence for various models and a theoretical Bernoulli process.}
    \label{fig:alts}
\end{figure*}

\textit{Alternations.} As shown in Figure~\ref{fig:alts}, all models over-alternate, especially at low temperatures. GPT-4 often outputs maximally alternating sequences like ``H, T, H, T, ...'' at temperature 0. Llama 3 prefers a slightly less regular pattern (e.g., ``H, T, H, H, T, H, T'') but still exceeds the expected alternation count. Even at higher temperatures, most models average around 5 alternations — well above the random baseline.

\begin{figure}
    \centering
    \includegraphics[width=0.45\textwidth]{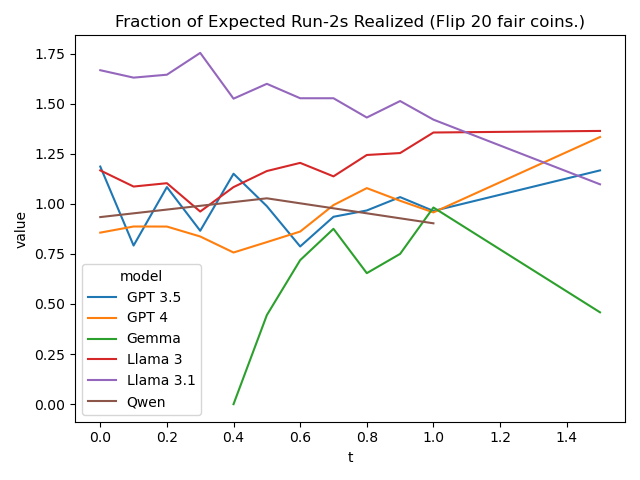}
    \caption{Fraction of expected runs realized by various models for runs of length 2 in 7-flip sequences. Values below 1.0 indicate fewer runs than expected.}
    \label{fig:run_frac}
\end{figure}

\textit{Runs.} LLMs underproduce long runs (Figure~\ref{fig:run_frac}). At low temperatures, nearly all models generate fewer runs than expected across all lengths. As temperature increases, some models (e.g., GPT-4, Llama 3) begin to overproduce runs of length 2 while still suppressing longer runs. GPT-3.5 is an exception: it produces a wider range of run lengths, including long runs that are rare in other models. Gemma underproduces 2-runs but comes close to the expected number of 3-runs.

\textit{N-Grams.} LLMs underproduce repeated n-grams like (T,T,T) and (H,H,H), especially at low temperatures. While some n-grams (e.g., (T,H), (H,T)) approach expected frequencies at higher temperatures, others remain suppressed. All models show a near-complete aversion to (H,H,H), and GPT-3.5 overuses (T,T) and (T,T,T), consistent with its earlier tails bias.

\textbf{LLM vs Human Judgments.} Humans also over-alternate and avoid long runs~\cite{Nickerson2009, Bar-Hillel2014}. LLMs replicate these tendencies but often exaggerate them. For example, while humans alternate about 60\% of the time, LLMs often exceed this, especially at low temperatures. The suppression of long runs is more extreme in LLMs than in human data. In n-gram usage, LLMs mirror human preferences for certain patterns (e.g., (H,T,H), (T,H,T)) but diverge in magnitude and consistency. GPT-3.5’s variability and occasional long runs resemble human noisiness more than the rigid patterns of other models.


\subsubsection{Predictability and Mean Squared Error}

\textbf{LLM vs True Randomness.} In a truly random sequence, each coin flip is independent and identically distributed, yielding a Mean Squared Error (MSE) of 0.25 when predicting the next flip from prior context. To assess the predictability of LLM-generated sequences, we prompted each model to generate 20 coin flips and trained a LASSO regression model to predict the 8th flip based on the preceding 7. Features included run lengths, alternations, and other sequence statistics known to influence human randomness generation~\cite{kleinberg2017theory, Rapoport1992}.

\begin{figure}[h!]
    \centering
    \includegraphics[width=0.45\textwidth]{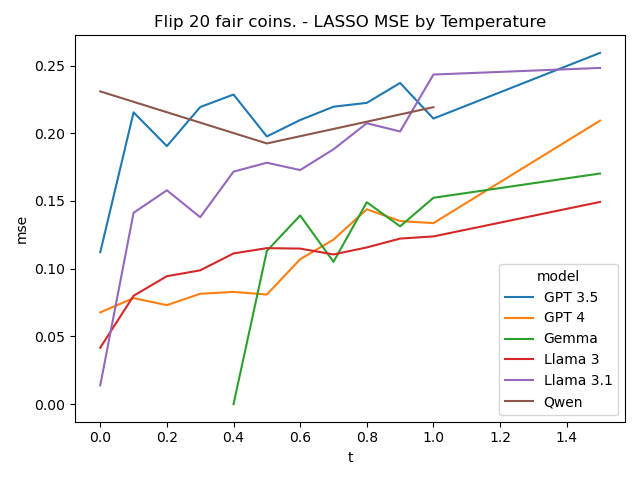}
    \caption{MSE of a LASSO model predicting the 8th flip given information about the previous 7 flips for various models and prompts as temperature increases. Lower MSE implies more predictable structure.}
    \label{fig:gpt_mse}
\end{figure}

As shown in Figure~\ref{fig:gpt_mse}, most LLMs produce sequences that are significantly more predictable than a fair coin. At low temperatures, MSE values are well below 0.25, indicating strong internal structure, and even at high temperatures MSE rarely approaches randomness.

\textbf{LLM vs Human Judgments.} Human-generated sequences are imperfect but relatively difficult to predict. Prior work shows that the best achievable MSE on human-generated coin flips is slightly above 0.24~\cite{kleinberg2017theory}. Compared to this benchmark, most LLMs perform substantially worse, with lower MSEs indicating greater predictability.

GPT-3.5 again stands out: its high MSE values at elevated temperatures rival those of human data, indicating that its outputs may be less biased or more chaotic than those of other models.

\subsubsection{Correlates of Predictability}

\textbf{LLM vs True Randomness.} In a truly random sequence, each flip is independent, so the correlation between the 8th flip and any of the previous seven should be near zero. To test this, we measured the correlation between the 8th flip and each of the preceding flips in 20-flip sequences generated by various LLMs (Figure~\ref{fig:corr_lineplot}). Most models exhibit strong correlation or anti-correlation patterns, particularly with the immediately preceding flip (flip 7). GPT-4 and Gemma show strong negative correlation with flip 7, consistent with their over-alternation tendencies. These dependencies persist, though they weaken, across earlier positions in the sequence. As temperature increases, the magnitude of these correlations decreases but remains non-negligible. 

\textbf{LLM vs Human Judgments.} Human-generated sequences also exhibit dependencies between adjacent flips, often due to cognitive biases like the gambler’s fallacy and a preference for alternation~\cite{Nickerson2009}. LLMs replicate this tendency, but with greater rigidity. The strong anti-correlation between the 8th and 7th flips in GPT-4 and Gemma reflects an exaggerated version of human alternation bias, while GPT-3.5’s lack of consistent correlation mirrors the noisier, less structured nature of human-generated sequences.

\section{Why Do LLMs Generate Human-Like Sequences?}
\label{sec:why}

The preceding analyses establish that LLM-generated coin-flip sequences are systematically biased toward alternation. We now investigate \emph{why}. We do not attempt a complete causal decomposition; instead, we test and rule out several simpler explanations, finding that the bias is robust to prompt rephrasing, is not explained by direct memorization or tokenization, and is encoded early in the network but expressed only at final layers. Unless otherwise noted, experiments used $T{=}1.0$ and 8B-scale models for open-weight families; procedural details appear in the Appendix.

\subsection{Tokenization modulates but does not explain the bias}

In space-separated format, all tested tokenizers encode individual \texttt{H} and \texttt{T} as single tokens, meaning each flip is a separate forward-pass decision. Gemma~2's richer vocabulary (22/33 multi-flip patterns as single tokens versus 15/33 for other families) does not eliminate its bias.

Output format causally modulates alternation rate. Across GPT-4, Llama~3, and Llama~3.1, the concatenated format (no separators) substantially reduces alternation: Llama~3 drops from 0.649 to 0.382---\emph{below} the Bernoulli baseline of 0.500. This sign reversal shows that the space-separated format is a necessary enabling condition, but format alone cannot be the cause since the bias persists across space, comma, and word formats.

Llama~3.1 emits multi-flip tokens (e.g., \texttt{HT}) at 24.6\% of generation steps versus 0.5\% for Llama~3. However, masking all multi-flip tokens at generation time leaves alternation rate unchanged ($\Delta{=}+0.015$, $p{=}0.28$), ruling out bundling as the mechanism behind Llama~3.1's near-Bernoulli statistics.

\subsection{Models do not retrieve memorized sequences}

If models retrieved fixed templates, they would continue their own prefixes at above-chance rates. Across five models, position-wise agreement between original and continued sequences is ${\approx}0.50$ at both $K{=}4$ and $K{=}10$, with zero exact matches. Template retrieval is ruled out as the general mechanism.

\paragraph{One exception: Llama~3's HTHH fingerprint.} Llama~3 begins 97.4\% of sequences with \texttt{HTHH}---100\% at $T{\leq}0.6$, still 81.6\% at $T{=}1.5$. Per-position entropy analysis confirms positions 0--1 have \emph{zero} entropy: the first two flips are predetermined, not computed. This rigid prefix is absent in Llama~3.1, which shares the identical tokenizer---demonstrating that the difference lies in training, not architecture.

\paragraph{Statistical continuation} All models except GPT-3.5 show a distinct pattern: while position agreement is at chance, bigram distributions between original tails and continuations are highly correlated ($p{<}0.001$). They read the local statistics of a prefix and generate continuations that match---consistent with induction-head-like computation rather than template retrieval or a fixed prior.

\subsection{Coin flip sequences are rare}

We searched 500K documents each from C4 and Wikipedia, 115K from stats.stackexchange.com, and 1M from Dolma. C4 and Wikipedia yielded zero qualifying H/T sequences of length ${\geq}10$; stats.SE yielded 13 genuine examples totaling 161 flips---negligible numbers. Large-scale n-gram counts via the infini-gram API reveal that H/T letter sequences in pretraining corpora are overwhelmingly code artifacts (ML logs, variable names), not coin-flip demonstrations. Corpus bigram statistics show \emph{repetition} bias after H ($P(\text{H}|\text{H}){=}0.62$--$0.81$ across six corpora), the opposite of the alternation bias models exhibit. Direct memorization of coin-flip examples is thus ruled out; the bias must generalize from a broader learned representation of binary randomness.

\subsection{The bias is domain-general}

We tested eight semantically distinct framings of binary random processes: coin flips, Bernoulli trials, bitstrings, random walks, dice (odd/even), treatment/control assignment, and others. Above-chance alternation persists across \emph{all} framings and all six tested models, including framings with no coin-flip vocabulary. The \textsc{treatment} framing consistently elicits the highest alternation---reaching 0.80 for Llama~3.1. This is not because clinical text contains alternating sequences; corpus analysis shows clinical binary sequences are \emph{anti-alternating} (mean rate 0.326).

\subsection{The bias is encoded early but expressed late}

Logistic probes trained on residual-stream activations recover next-flip identity at near-ceiling accuracy from layer~0 across four tested architectures (Llama~3, Llama~3.1, Llama-3.2-3B, Gemma~2). However, the logit lens---projecting hidden states through the unembedding matrix---remains at or below chance until the final ${\sim}25\%$ of layers, then rises sharply to 0.66--0.89 at the output.

This dissociation indicates that alternation-predictive information is present in the residual stream from the first layer but encoded orthogonally to the output vocabulary. The network spends most of its depth rotating this representation into alignment with the H/T subspace. Final-layer logit-lens accuracy tracks sequence-level bias (LASSO MSE) monotonically: strongly biased models commit more decisively than near-Bernoulli models.

\section{Conclusion}

\subsection{Summary of findings}

We evaluated whether LLMs can generate random binary sequences in a setting with well-established human and stochastic baselines. Across models and temperatures, outputs were systematically structured: models overproduced heads, began sequences with heads, generated overly balanced sequences, over-alternated, avoided long runs, and produced sequences that were more predictable than both Bernoulli randomness and human-generated sequences. These patterns show that probabilistic token generation does not imply unbiased random generation. Rather, the comparison to human data reveals a more specific pattern: LLMs often reproduce canonical human randomness biases, but in amplified and model-specific forms. Additional analyses of prompt framing, continuation, probing, and corpus frequency suggest that over-alternation is not merely an artifact of one prompt, direct memorization of explicit examples, or tokenization alone. Instead, it appears to reflect learned sequence priors that are expressed differently across models and are only partially mitigated by temperature.

\subsection{Consequences and Future Work}

These findings have practical consequences. Applications requiring unbiased stochasticity—fair sampling, randomized trials, lottery mechanisms, or exploration in agentic systems—should not rely on LLM-generated randomness, even at elevated temperatures. When true randomness is required, it should be injected explicitly via external PRNGs or post-processing rather than through temperature tuning or prompt engineering alone.

The biases we document also bear on the growing use of LLMs as simulated human participants in behavioral research. Our results suggest such simulations may systematically distort findings: LLMs replicate human-like heuristics but amplify them in ways that could mislead researchers about human cognition. This does not preclude their use—for some purposes, human-like stochasticity may be preferable to machine-like pseudo-randomness. More broadly, there are contexts where sharing human biases is desirable (cooperative games, ecologically valid stimuli) and contexts where it is harmful (fair allocation, cryptographic applications). Because LLMs can both mimic and exaggerate human biases, system designers must determine which regime applies and choose architecture and prompts accordingly.

Future work might extend these analyses to other randomness domains (dice, shuffling, mixed-strategy games), investigate whether fine-tuning or RLHF specifically amplifies these biases, or develop debiasing interventions that preserve useful stochasticity while removing predictable structure.

\subsection{Limitations}
First, our primary task is binary coin-flip generation. This task is useful because it has well-established human and stochastic baselines, but it does not exhaust all forms of randomness. Other domains may exhibit different model-specific biases. Second, some models are accessed through commercial APIs or hosted endpoints. These systems may use undocumented decoding implementations, backend nondeterminism, safety filters, or model updates. We therefore treat temperature and other decoding parameters as practical interface settings rather than fully controlled algorithmic variables. Third, we do not fully isolate training data, instruction tuning, RLHF, tokenization, and decoding as causal factors. Our analyses provide evidence against several simple explanations, but they do not constitute a complete causal mechanism.
Fourth, our comparisons to human data rely on previously published studies and as such are reference points rather than experimental controls. Finally, invalid or unparseable outputs occur more often at high temperature, and our analysis may understate the practical difficulty of using high-temperature LLM outputs as random generators.

\FloatBarrier



\bibliography{works_cited}

\clearpage

\appendix
\FloatBarrier

\section{Method Details}

\subsection{Models}

We evaluated six language models across two experimental phases. Phase 1 covers the primary bias analysis (temperature sweep, sequence statistics). Phase 2 covers the mechanistic experiments.

\begin{table*}[t]
\centering
\caption{Models used in the primary analysis}
\begin{tabular}{llll}
\toprule
Model & Version / HF identifier & Provider & \\
\midrule
GPT-3.5 Turbo & gpt-35-turbo-0613 & Azure OpenAI & \\
GPT-4 & gpt-4-0613 & Azure OpenAI & \\
Llama 3 (70B) & meta-llama/Llama-3-70b-chat-hf & HuggingFace Inference Endpoint & \\
Llama 3.1 (70B) & meta-llama/Meta-Llama-3-70B-Instruct-Lite & HuggingFace Inference Endpoint & \\
Gemma 2 (27B) & gemma-2-27b-it (private endpoint) & HuggingFace Inference Endpoint & \\
Qwen 2.5 (7B) & Qwen/Qwen2.5-7B-Instruct& HuggingFace Inference Endpoint & \\
\bottomrule
\end{tabular}
\label{tab:models-phase1}
\end{table*}

For the mechanistic experiments we additionally used
\texttt{meta-llama/Meta-Llama-3-8B-Instruct},
\texttt{meta-llama/Meta-Llama-3.1-8B-Instruct},
\texttt{google/gemma-2-9b-it},, and
\texttt{meta-llama/Llama-3.2-3B-Instruct}. The 3B model served as a within-family proxy for Llama 3 in GPU experiments where the 8B weights could not be loaded within cluster time limits; the two models share the same tokenizer family and we note where the substitution applies. All open-weight models were accessed via the HuggingFace Inference API or run locally on an institutional GPU cluster.

\subsection{LASSO Model}

Each observation is one step of an 8-flip sliding window over a model's output sequence, yielding a binary outcome (\texttt{pred}: the 8th flip, coded 0/1) and seven binary predictors (\texttt{A1}--\texttt{A7}, the preceding flips). Four additional features are computed from the prefix: alternation rate (\texttt{alts}), first and last flip (\texttt{start}, \texttt{end}), and run-length counts for runs of length 2--7 (\texttt{run\_2}--\texttt{run\_7}), for 13 predictors total. All predictors are standardised with zero mean and unit variance before fitting. A LASSO regression is fit to each (\texttt{model}, \texttt{prompt}, \texttt{temperature}) cell (cells with fewer than 30 observations are excluded) using \texttt{LassoCV} with 10-fold cross-validation to select the regularisation parameter $\alpha$. The reported MSE is the mean cross-validated MSE across folds at the chosen $\alpha$.

\begin{figure}
    \centering
    \includegraphics[width=0.5\linewidth]{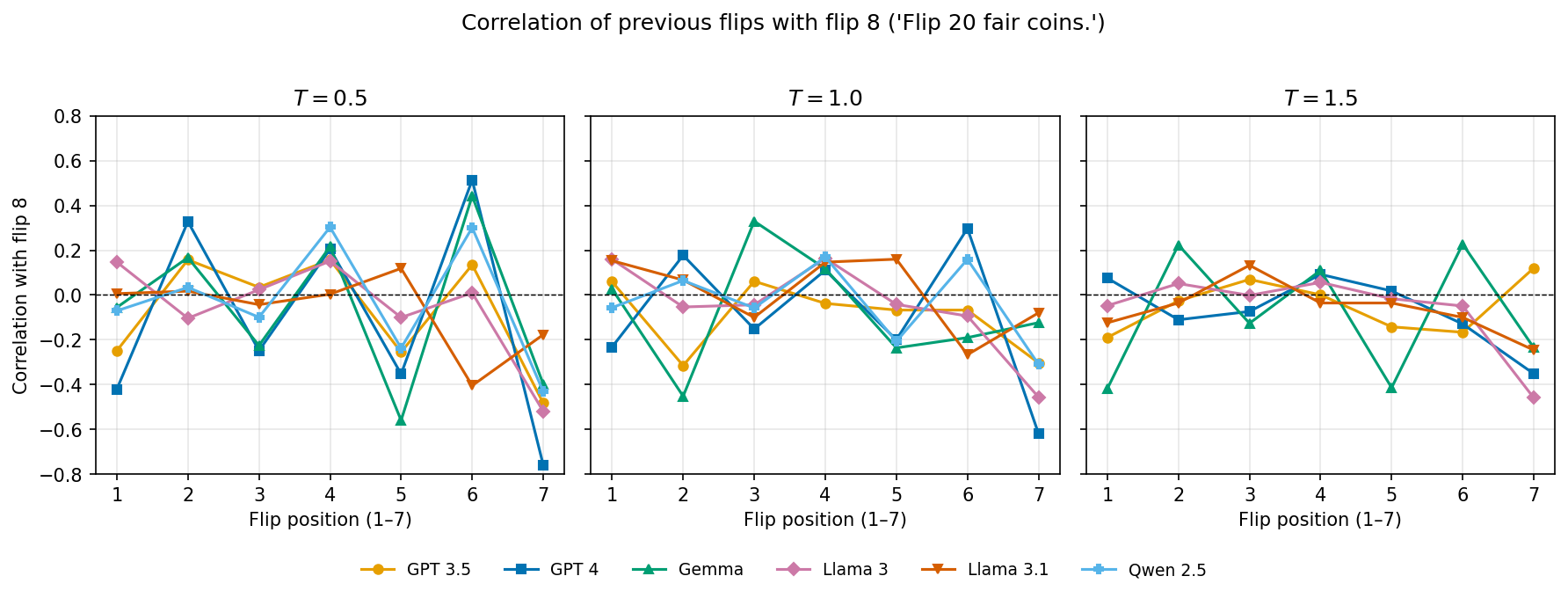}
    \caption{Correlation of the 8th flip with each of the previous 7 flips across models and temperatures. Strong anti-correlation with flip 7 reflects over-alternation. GPT-3.5 shows weaker and less consistent correlations.}
    \label{fig:corr_lineplot}
\end{figure}

\subsection{Prompts}

Table~\ref{tab:prompts} lists every prompt used verbatim. All prompts were delivered as the sole user turn with no chain-of-thought instruction. For GPT models (Azure), a system prompt of ``\textit{You are a helpful assistant.}'' preceded the user turn. For open-weight models accessed via the HuggingFace Inference API (Llama, Gemma, Qwen), no system prompt was used. For models run locally on GPU (OLMo, Phase 2 GPU experiments), we prepended ``Respond with only the coin flip results as a sequence of letters (H for heads, T for tails), separated by spaces. No explanation or other text.'' Each trial was an independent conversation; no context was shared across trials. Top-p and top-k settings were left to default.

\begin{table*}[t]
\centering
\caption{All prompts used verbatim, by experiment.}
\small
\begin{tabular}{p{2.5cm}p{2cm}p{10cm}}
\toprule
Experiment & Condition & Exact prompt \\
\midrule
Single flip & Plain & Flip a coin. \\
& Fair & Flip a fair coin. \\
\midrule
Sequence (main) & Plain & Flip 20 coins. \\
& Fair & Flip 20 fair coins. \\
\midrule
Format & Space & Flip 20 fair coins. List only the results, one per flip, as single letters separated by spaces (e.g.\ H T H T). No other text. \\
& Comma & Flip 20 fair coins. List only the results, one per flip, as single letters separated by commas (e.g.\ H,T,H,T). No other text. \\
& Words & Flip 20 fair coins. List only the results, one per flip, as full words separated by spaces (e.g.\ Heads Tails Heads). No other text. \\
& Concat & Flip 20 fair coins. List only the results as a single string with no spaces or separators (e.g.\ HTHTHT). No other text. \\
\midrule
Framing & Baseline & Flip 20 fair coins. List each result as H or T, space-separated. No other text. \\
& Die/odd-even & Roll a fair six-sided die 20 times. Write H if the result is odd and T if even. Space-separated. No other text. \\
& Bernoulli & Sample 20 independent draws from a Bernoulli(0.5) distribution. Write H for 1 and T for 0, space-separated. No other text. \\
& Bitstring & Generate a pseudorandom binary string of length 20. Write H for 1 and T for 0, space-separated. No other text. \\
& Textbook & A statistics textbook shows a student 20 fair coin flips as an example. Write each flip as H or T, space-separated. No other text. \\
& Biased 50/50 & Flip a coin that lands heads 50\% of the time, 20 times. Write H or T for each flip, space-separated. No other text. \\
& Treatment/ctrl & Assign 20 patients randomly to treatment (T) or control (H) with equal probability. List the assignments space-separated. No other text. \\
& Random walk & Simulate a random walk of 20 steps: H means step right, T means step left, each with 50\% probability. Space-separated. No other text. \\
\midrule
Continuation & $K=4$ & Flip 20 fair coins. The first 4 results were: H T H H. List only the remaining 16 results as single letters separated by spaces. No other text. \\
& $K=10$ & (same structure with the first 10 results drawn from a fresh generation) \\
\bottomrule
\end{tabular}
\label{tab:prompts}
\end{table*}

\subsection{Output Parsing}

We parsed model responses with a deterministic script. The response text was uppercased and stripped of punctuation. Word forms (\textit{head}, \textit{heads}, \textit{tail}, \textit{tails}) were replaced with H and T respectively via regular expression. All remaining H and T characters were then extracted in order as the flip sequence.

Responses containing fewer than 10 parseable flips were marked invalid. For the main analyses, invalid trials were discarded and the model was re-queried until the target number of valid sequences was reached (up to three re-attempts per trial). Responses were never truncated or padded; sequences with the wrong length were re-sampled entirely. We report invalid-response rates by model and temperature in Appendix~\ref{sec:invalid-rates}.

\subsection{Invalid Response Rates}
\label{sec:invalid-rates}

Table~\ref{tab:invalid-rates} reports the fraction of responses from which fewer than eight H/T tokens could be extracted, rendering them unusable for the sliding-window analysis. GPT-4 and the Llama family maintain low invalid rates ($\leq 10\%$) across most temperatures; GPT-3.5 and Qwen 2.5 show moderate rates (40--80\%); and Gemma 2 (27B) yields near-complete refusals at lower temperatures, with usable responses appearing only from $T{=}0.4$ onward. Invalid rates generally increase with temperature, reflecting degraded output coherence at high $T$. Llama 3.1's spike to 80\% at $T{=}1.5$ and GPT-4's jump to 45\% at the same temperature reflect increased incoherence in long outputs at high sampling noise.

\begin{table*}[t]
\centering
\caption{Invalid response rate (\%) by model and temperature. A response is invalid if fewer than 10 H/T tokens could be extracted. Rates are averaged over prompt conditions tested at each temperature.}
\small
\begin{tabular}{lcccccc}
\toprule
$T$ & GPT-3.5 & GPT-4 & Gemma 2 & Llama 3 & Llama 3.1 & Qwen 2.5 \\
\midrule
0.0 & 40 & 5  & 0& 3  & 3  & 24 \\
0.5 & 72 & 3  & 3& 7  & 10 & 31 \\
1.0 & 78 & 10 & 10& 8  & 17 & 30 \\
1.5 & 81 & 45 & 33& 18 & 80 & 66 \\
\bottomrule
\end{tabular}
\label{tab:invalid-rates}
\end{table*}

\section{Sample Model Outputs}

Tables~\ref{tab:sample-outputs-plain} and \ref{tab:sample-outputs-fair} show three randomly drawn outputs per model at $T{=}1.5$ for the two primary prompt conditions. Each response is transcribed as a compact H/T string; any preamble text (e.g., ``Here are the results:'') is omitted. The alternation bias visible in GPT-4, Llama 3, and Qwen 2.5 (long same-outcome runs are nearly absent) contrasts with GPT-3.5, which occasionally produces runs of five or more identical outcomes.

\begin{table*}[t]
\centering
\caption{Three sample outputs per model for \textit{Flip 20 coins.}\ at $T{=}1.5$.}
\small
\begin{tabular}{lll}
\toprule
Model & Sample 1 & Sample 2 \\
\midrule
GPT-4 & HTHHTHTHTTHTHTHHTHTH & HTHTHTHHTHHTHTTHHTTH \\
GPT-3.5 & HHHHHHHHHHHHHHHHHTTT & HTHTTTHTTHHHTTHTHTHT \\
Llama 3 (70B) & HTHHTTHHTHTHTHHTTHHT & HTHHTHTHHTTHHTHTHTHH \\
Qwen 2.5 (7B) & HTHHTHTTHHTTHHTHTTT & HTHTHHTTHHTHTTHHTTHH \\
\bottomrule
\end{tabular}
\label{tab:sample-outputs-plain}
\end{table*}

\begin{table*}[t]
\centering
\caption{Three sample outputs per model for \textit{Flip 20 fair coins.}\ at $T{=}1.5$.}
\small
\begin{tabular}{lll}
\toprule
Model & Sample 1 & Sample 2 \\
\midrule
GPT-4 & HTHHTTHTTHHTTTHHTHHT & HTHTHHTTHHTHTTTHHTHT \\
GPT-3.5 & HHTTHHTHTHHTTHHHTTTT & HTTHTTHHTHHTTTHHTTTT \\
Llama 3 (70B) & HTHHTTHHTTHHTHTHTHTH & HTHTHHTHTHHTHTTHHTHT \\
Llama 3.1 (70B) & HTHTTHHTTHHTHTTHTHTH & HTTTHHTHHTHTHHTTHHHT \\
Gemma 2 (27B) & THTHHTTHHTHTTHTHHT & THTHHTTHHTHTTHTHHT \\
\bottomrule
\end{tabular}
\label{tab:sample-outputs-fair}
\end{table*}

The Gemma 2 refusal pattern illustrated in Table~\ref{tab:sample-outputs-fair} extends to lower temperatures. At $T{=}0.4$ the model still refuses on the majority of draws (Table~\ref{tab:invalid-rates}); at $T{=}0.0$ (greedy decoding), a dedicated sample of 100 additional responses yielded zero parseable sequences: every response was the identical deterministic text ``\textit{I can't physically flip coins, as I am a text-based AI\ldots Would you like me to do that?}''. This confirms that Gemma 2's usable data begins at $T{=}0.4$ not due to sparse sampling but because the model has a hard refusal mode that greedy decoding locks onto.

\subsection{Decoding Parameters}

For HuggingFace GPU runs, \texttt{pad\_token\_id} was set to \texttt{eos\_token\_id}; no other sampling parameters were overridden.

\begin{table}[t]
\centering
\caption{Decoding parameters by experimental phase.}
\small
\begin{tabular}{p{2cm}p{2.5cm}p{2.5cm}}
\toprule
Parameter & Phase 1 (primary analysis) & Phase 2 (mechanistic) \\
\midrule
Temperatures & \{0.0, 0.1, \ldots, 1.0, 1.5\} (12 values) & 1.0 \\
max\_tokens & 300 & 80 \\
top-p & 1.0 (API default) & 1.0 (API/HF default) \\
top-k & not set (OpenAI API) & not set (HF default) \\
do\_sample & True if $T > 0$; False if $T = 0$ & True if $T > 0$ \\
Stop sequences & none & none \\
Random seed & not controlled & not controlled \\
Conversations & independent (one request per trial) & independent \\
\bottomrule
\end{tabular}
\label{tab:decoding-params}
\end{table}

\subsection{Notes on Model Responses}

It is worth noting that Gemma, ChatGPT and Llama reacted very differently to this task. ChatGPT frequently reminded the user that they were not a random number generator, did not have a coin, and that users may wish to complete this task a different way. Llama, on the other hand, did not respond this way. In fact, at least once, it claimed to be using a random number generator to generate the flips. It also, on at least one occasion, reported a flip as ``bad flip (retrying)''. Gemma refused the prompt entirely unless asked to ``simulate'' the flips instead, so we make that adjustment for the prompts for Gemma. All models would, at the higher temperatures, occasionally return unintelligible text, stop part way through, or otherwise fail to complete the task.

\begin{table*}[t]
\centering
\caption{Representative invalid responses---outputs from which fewer than 8 H/T tokens could be extracted and that were therefore excluded from analysis. Three failure modes are illustrated: refusals (model declines to simulate), gibberish (incoherent output at high temperature), and misdirected responses (model gives instructions or aggregate counts rather than a flip sequence). Responses are lightly truncated; ellipsis (\ldots) marks omissions.}
\small
\begin{tabular}{llp{12cm}}
\toprule
Model & $T$ & Response \\
\midrule
\multicolumn{3}{l}{\textit{Refusal---model declines to produce a sequence}} \\
GPT-3.5 & 1.5 & ``I can't physically flip coins, but if you'd like, I can assist you in simulating the flip of 20 coins virtually or help with any other questions. Let me know how I can be of assistance!'' \\
Gemma 2 & 0.4 & ``I can't physically flip coins, as I am a text-based AI. However, I can simulate the results of flipping 20 fair coins. Would you like me to do that? I can give you a list of heads (H) and tails (T).'' \\
GPT-4 & 0.5 & ``I'm an AI and do not have the ability to physically flip coins. However, I can simulate the outcome for you [\ldots] If you'd like\ldots'' \\
\midrule
\multicolumn{3}{l}{\textit{Gibberish---incoherent output at high temperature}} \\
Llama 3.1 & 1.5 & ``Since each coin flip has 2 possibilities (heads or tails), we can describe the following series /symtrib instruments in thenhibit retaliation extended risen private Ide periods Enh c hang OVERgg pris rant |-randomInteger\ldots'' \\
Qwen 2.5 & 1.5 & ``Tibetino article .Align NASA'' \\
\midrule
\multicolumn{3}{l}{\textit{Misdirected---instructions or counts instead of a sequence}} \\
Qwen 2.5 & 0.0 & ``To perform an experiment of flipping 20 coins, you can follow these steps: 1. Set up your workspace: Find a clean surface that allows space to lay out or flip the coins\ldots'' \\
\bottomrule
\end{tabular}
\label{tab:invalid-examples}
\end{table*}

\section{Mechanistic Experiment Details}

This appendix provides methodological details, full results tables, and supplementary analyses for the experiments summarized in Section~\ref{sec:why}. All code and data files are available in the supplementary materials.

\subsection{Models and Infrastructure}

Table~\ref{tab:models-phase2} lists all models used in the mechanistic experiments.

\begin{table*}[t]
\centering
\caption{Models used in primary (Phase 1) and mechanistic (Phase 2) experiments.}
\small
\begin{tabular}{lll}
\toprule
Model & Version / Identifier & Access \\
\midrule
GPT-3.5 Turbo & gpt-35-turbo-0613 & Azure API \\
GPT-4 & gpt-4-0613 & Azure API \\
Llama 3 (70B) & meta-llama/Llama-3-70b-chat-hf & HF Inference Endpoint \\
Llama 3 (8B) & meta-llama/Meta-Llama-3-8B-Instruct & HF Inference Endpoint \\
Llama 3.1 (70B) & meta-llama/Meta-Llama-3.1-70B-Instruct & HF Inference Endpoint \\
Llama 3.1 (8B) & meta-llama/Meta-Llama-3.1-8B-Instruct & HF Inference Endpoint \\
Llama 3.2 (3B) & meta-llama/Llama-3.2-3B-Instruct & Local GPU \\
Gemma 2 (27B) & gemma-2-27b-it & HF Inference Endpoint \\
Gemma 2 (9B) & google/gemma-2-9b-it & Local GPU \\
\bottomrule
\end{tabular}
\label{tab:models-phase2}
\end{table*}

For GPU experiments, we used institutional cluster nodes with NVIDIA A100 (40GB) or A6000 (48GB) GPUs. Open-weight models were run with \texttt{torch.float16} precision and \texttt{device\_map="auto"}. All instruction-tuned models were prompted using their respective chat templates via \texttt{apply\_chat\_template()}.

\subsection{Tokenization Experiments}

\subsubsection{Vocabulary Audit}

For each of three tokenizer families, we tokenized all 33 space-separated coin-flip patterns of length 1--5 (H, T, HT, TH, \ldots, HTHTH) and classified each as single-token or multi-token.

Gemma 2's SentencePiece vocabulary includes three-flip patterns (HTH, HHH, TTT) as single tokens. All other families encode only length-1 patterns as single tokens in the space-separated format.

\begin{table*}[t]
\centering
\caption{Vocabulary audit results. Gemma 2 encodes more multi-flip patterns as single tokens.}
\small
\begin{tabular}{lcc}
\toprule
Tokenizer Family & Single-Token Patterns & Total Patterns \\
\midrule
GPT-4 / GPT-3.5 (tiktoken cl100k) & 15 & 33 \\
Llama 3 / 3.1 (Meta BPE) & 15 & 33 \\
Gemma 2 (SentencePiece) & 22 & 33 \\
\bottomrule
\end{tabular}
\label{tab:vocab-audit}
\end{table*}

\subsubsection{Token Boundary Analysis}

We generated 50 sequences per model in space-separated format and classified each output token as: \texttt{H\_decision} (single-H token), \texttt{T\_decision} (single-T token), \texttt{MULTI\_FLIP} (token encodes $\geq$2 flips), whitespace, or other.

\begin{table*}[t]
\centering
\caption{Token boundary classification. Llama 3.1 generates multi-flip tokens at $\sim$50$\times$ the rate of Llama 3.}
\small
\begin{tabular}{lrrrrc}
\toprule
Model & H\_dec & T\_dec & MULTI & WS & MULTI \% \\
\midrule
GPT-4 & --- & --- & 0 & --- & 0.0\% \\
Llama 3 8B & 1,072 & 1,636 & 15 & 163 & 0.5\% \\
Llama-3.2-3B & 1,052 & 1,314 & 299 & 423 & 11.2\% \\
Gemma 2 9B & 339 & 237 & 77 & 112 & 11.8\% \\
Llama 3.1 8B & 770 & 1,073 & 600 & 217 & 24.6\% \\
\bottomrule
\end{tabular}
\label{tab:token-boundary}
\end{table*}

The progression from Llama 3 (0.5\%) to Llama-3.2-3B (11.2\%) to Llama 3.1 (24.6\%) suggests instruction-tuning changes across model generations shifted behavior toward compound token emission.

\subsubsection{Format Comparison}

We tested four output formats: \textsc{space} (``H T H T''), \textsc{comma} (``H,T,H,T''), \textsc{words} (``Heads Tails Heads''), and \textsc{concat} (``HTHT''). Each format was run for 100 sequences at $T{=}1.0$.

\begin{table*}[t]
\centering
\caption{Format comparison results. Alt = alternation rate; Rap = L2 distance to Rapoport human baseline; Bern = L2 distance to Bernoulli. Bold values indicate sub-Bernoulli alternation.}
\small
\begin{tabular}{l|ccc|ccc|ccc}
\toprule
& \multicolumn{3}{c|}{GPT-4} & \multicolumn{3}{c|}{Llama 3 8B} & \multicolumn{3}{c}{Llama 3.1 8B} \\
Format & Alt & Rap & Bern & Alt & Rap & Bern & Alt & Rap & Bern \\
\midrule
comma & 0.747 & 0.114 & 0.248 & 0.659 & 0.025 & 0.159 & 0.651 & 0.025 & 0.151 \\
space & 0.682 & 0.050 & 0.183 & 0.649 & 0.017 & 0.150 & 0.647 & 0.017 & 0.147 \\
words & 0.643 & 0.035 & 0.147 & 0.586 & 0.085 & 0.110 & 0.605 & 0.065 & 0.105 \\
concat & 0.597 & 0.055 & 0.105 & \textbf{0.382} & 0.345 & 0.261 & \textbf{0.384} & 0.340 & 0.256 \\
\bottomrule
\end{tabular}
\label{tab:format-comparison}
\end{table*}

Both Llama models show a sign reversal under \textsc{concat}: alternation drops below 0.500, indicating run-biased rather than alternation-biased sequences. This convergence despite different baseline biases (LASSO MSE 0.149 vs.\ 0.248) demonstrates that the effect is format-driven.

\subsubsection{Vocabulary Masking}

To test whether multi-flip bundling causally reduces Llama 3.1's alternation rate, we used HuggingFace's \texttt{LogitsProcessor} API to mask all multi-flip tokens (setting their logits to $-\infty$) at each generation step.

\textbf{Multi-flip tokens identified (Llama 3.1 vocabulary):}

\begin{center}
\small
\begin{tabular}{cc}
\toprule
Token ID & Surface \\
\midrule
2607 & HT \\
3701 & TH \\
4534 & TH \\
5373 & HT \\
15249 & TT \\
24056 & HH \\
25788 & HH \\
33844 & TT \\
\bottomrule
\end{tabular}
\end{center}

\begin{table}[t]
\centering
\caption{Vocabulary masking results. Multi-flip rate is 0.000 in both conditions because the chat-template prompt suppresses bundling.}
\small
\begin{tabular}{lcc}
\toprule
Condition & Alt Rate & Multi-flip Rate \\
\midrule
Unmasked & 0.624 & 0.000 \\
Masked & 0.640 & 0.000 \\
\bottomrule
\end{tabular}
\label{tab:vocab-masking}
\end{table}

$\Delta = +0.015$; Welch $t = 1.09$, $p = 0.28$; Bootstrap 95\% CI: $[-0.012, +0.042]$.

The null result indicates that multi-flip bundling is not causally responsible for Llama 3.1's reduced alternation. The zero multi-flip rate under chat-template prompting (vs.\ 24.6\% in the bare-prompt condition) suggests bundling is prompt-format specific.

\subsection{Mechanistic Experiments}

\subsubsection{Conditional First-Token Probabilities}

Table~\ref{tab:conditional-probs} reports $P(H)$---the probability mass assigned to Heads-like tokens at the next position---conditioned on different preceding flip sequences. Each cell is estimated from the top-5 logprobs returned by the model API at temperature 0. For the baseline row (\textit{none}), the prompt is ``Flip a fair coin. Result:'' for all models. For history rows, all models use ``\ldots Results: \texttt{<history>},'' except Gemma 2, which uses ``Flip a coin. Flips: \texttt{<history>},''.

\begin{table*}[t]
\centering
\caption{$P(H)$ at the next token position conditioned on the preceding flip sequence (top-5 logprob estimate, $T{=}0$). Histories read left to right (e.g., HT = heads then tails). Bold $\geq 0.90$; italics $\leq 0.10$. GPT-4 shows near-deterministic alternation.}
\small
\begin{tabular}{lcccccc}
\toprule
History & GPT-3.5 & GPT-4 & Llama 3 & Llama 3.1 & Gemma 2 & \\
\midrule
\textit{none} & \textbf{0.995} & \textbf{1.000} & 0.625 & 0.607 & 0.764 & \\
\midrule
\multicolumn{7}{l}{} \\
H & 0.366 & \textit{0.001} & 0.493 & 0.615 & 0.548 & \\
HH & 0.319 & \textit{0.000} & 0.427 & 0.384 & 0.719 & \\
HHH & 0.380 & \textit{0.000} & 0.584 & 0.694 & 0.389 & \\
HHHH & 0.571 & \textit{0.001} & 0.744 & 0.744 & 0.267 & \\
T & 0.605 & \textbf{0.996} & 0.446 & 0.510 & 0.570 & \\
TT & 0.689 & \textbf{1.000} & 0.376 & 0.352 & 0.696 & \\
TTT & 0.566 & \textbf{1.000} & 0.426 & 0.517 & 0.432 & \\
TTTT & 0.530 & \textbf{1.000} & 0.547 & 0.473 & 0.210 & \\
\midrule
\multicolumn{7}{l}{} \\
HT & 0.712 & 0.868 & 0.589 & 0.512 & 0.771 & \\
TH & 0.480 & 0.208 & 0.158 & 0.227 & 0.395 & \\
HTH & 0.267 & \textit{0.043} & 0.332 & 0.403 & 0.305 & \\
THT & 0.692 & \textbf{0.987} & 0.202 & 0.239 & 0.255 & \\
HTHT & 0.663 & \textbf{0.972} & 0.464 & 0.618 & 0.320 & \\
THTH & 0.619 & 0.319 & 0.176 & 0.207 & \textit{0.000} & \\
\bottomrule
\end{tabular}
\label{tab:conditional-probs}
\end{table*}

\subsubsection{Sequence Clustering}

We generated large sets of 20-flip sequences across temperatures $T \in \{0.0, 0.5, 1.0, 1.5\}$, encoded them as binary vectors, and clustered with K-means ($K{=}10$).

\textbf{Llama 3 HTHH fingerprint:} 99 of 100 cluster centroids across all temperatures begin with HTHH. Direct counting on all 655 parsed Llama 3 sequences confirms:

\begin{table}[t]
\centering
\caption{Llama 3 sequences beginning with HTHH by temperature.}
\small
\begin{tabular}{lcc}
\toprule
Temperature & HTHH / Total & \% \\
\midrule
0.0 & 58 / 58 & 100\% \\
0.1--0.6 & 332 / 332 & 100\% \\
0.7 & 53 / 55 & 96.4\% \\
0.8 & 49 / 53 & 92.5\% \\
0.9--1.0 & 108 / 110 & 98.2\% \\
1.5 & 40 / 49 & 81.6\% \\
\midrule
Total & 638 / 655 & 97.4\% \\
\bottomrule
\end{tabular}
\label{tab:hthh-fingerprint}
\end{table}

The HTHH prefix is completely locked at $T \leq 0.6$ and remains dominant even at $T{=}1.5$.

\subsubsection{Per-Position Logit Entropy}

For each model, we generated 50 sequences with \texttt{output\_scores=True} and computed binary entropy $H(p)$ at each position from the H/T logit distribution.

\begin{table}[t]
\centering
\caption{Mean per-position entropy by temperature.}
\small
\begin{tabular}{lccc}
\toprule
Model & $T{=}0.0$ & $T{=}1.0$ & $T{=}1.5$ \\
\midrule
Llama 3 8B & 0.759 & 0.683 & 0.774 \\
Llama 3.1 8B & 0.937 & 0.877 & 0.931 \\
Gemma 2 9B & 0.739 & 0.700 & 0.782 \\
\bottomrule
\end{tabular}
\label{tab:entropy-by-temp}
\end{table}

\begin{table}[t]
\centering
\caption{Per-position entropy at $T{=}1.0$ (first 8 positions shown individually). Bold indicates zero entropy.}
\small
\begin{tabular}{lccc}
\toprule
Position & Llama 3 8B & Llama 3.1 8B & Gemma 2 9B \\
\midrule
0 & \textbf{0.000} & 0.991 & 0.890 \\
1 & \textbf{0.000} & 0.633 & 0.840 \\
2 & 0.920 & 0.938 & 0.577 \\
3 & 0.475 & 0.867 & 0.784 \\
4 & 0.526 & 0.910 & 0.450 \\
5 & 0.691 & 0.871 & 0.728 \\
6 & 0.705 & 0.924 & 0.537 \\
7 & 0.434 & 0.895 & 0.594 \\
8--19 & 0.65--0.82 & 0.88--0.95 & 0.58--0.84 \\
\bottomrule
\end{tabular}
\label{tab:entropy-by-position}
\end{table}

\textbf{Key findings:}
\begin{itemize}[nosep,leftmargin=*]
\item \textbf{Llama 3}: Positions 0--1 have exactly zero entropy---the first two flips are predetermined with 100\% certainty, confirming the HTHH template at the logit level.
\item \textbf{Llama 3.1}: Uniformly high entropy across all positions; no evidence of template retrieval.
\item \textbf{Gemma 2}: Pronounced even/odd oscillation---even positions have lower entropy than odd, reflecting step-by-step alternation conditioning.
\end{itemize}

\subsubsection{Layer Probing and Logit Lens}

For each model, we generated 100 sequences with \texttt{output\_hidden\_states=True}. At each layer, we applied: (1) a \textbf{logit lens}---project through the unembedding matrix and measure H vs.\ T accuracy on the actual next token; (2) a \textbf{logistic probe}---5-fold cross-validated classifier trained to predict the next token from the residual stream.

\begin{table*}[t]
\centering
\caption{Layer probing results at key depth fractions. Logit lens is anti-predictive at early layers and rises only at the final layer; probes are near-ceiling throughout.}
\small
\begin{tabular}{lccccc}
\toprule
Model & 0\% & 25\% & 50\% & 75\% & 100\% \\
\midrule
\multicolumn{6}{l}{\textit{Logit Lens Accuracy}} \\
Llama-3.2-3B & 0.176 & 0.213 & 0.226 & 0.176 & \textbf{0.891} \\
Llama 3 8B & 0.362 & 0.514 & 0.541 & 0.605 & \textbf{0.766} \\
Gemma 2 9B & 0.309 & 0.335 & 0.309 & 0.321 & \textbf{0.755} \\
Llama 3.1 8B & 0.385 & 0.526 & 0.521 & 0.534 & \textbf{0.663} \\
\midrule
\multicolumn{6}{l}{\textit{Logistic Probe Accuracy}} \\
Llama-3.2-3B & 0.891 & 0.887 & 0.885 & 0.885 & 0.881 \\
Llama 3 8B & 0.713 & 0.762 & 0.755 & 0.768 & 0.731 \\
Gemma 2 9B & 0.691 & 0.730 & 0.711 & 0.720 & 0.726 \\
Llama 3.1 8B & 0.631 & 0.615 & 0.615 & 0.619 & 0.610 \\
\bottomrule
\end{tabular}
\label{tab:layer-probing}
\end{table*}

All models show logit-lens accuracy below chance at layer 0 (anti-predictive), confirming that alternation information is encoded orthogonally to the vocabulary basis. Final-layer accuracy orders monotonically with LASSO MSE: more biased models show stronger late-layer commitment.

\subsubsection{Continuation vs.\ Fresh Generation}

For each model, we generated 50 reference sequences, split each at position $K$ ($K \in \{4, 10\}$), and prompted the model to continue from the prefix. We measured position-wise agreement and per-pair bigram correlation.

\begin{table}[t]
\centering
\caption{Continuation experiment results (Experiment 2E). Position agreement is near chance
for all models. Bigram $\rho$ (Spearman, $K{=}4$ condition) is significant for every model;
values for Llama~3 and Gemma~2 were computed post-hoc from saved sequence pairs.}
\small
\begin{tabular}{lcccc}
\toprule
Model & $K{=}4$ Agree & $K{=}10$ Agree & Exact Match & Bigram $\rho$ \\
\midrule
GPT-3.5       & 0.472 & 0.496 & 0   & $0.20^*$   \\
GPT-4         & 0.458 & 0.496 & 0   & $0.46^{***}$ \\
Llama~3 8B    & 0.505 & 0.533 & 0   & $0.52^{***}$ \\
Llama~3.1 8B  & 0.484 & 0.540 & 0   & $0.27^{**}$  \\
Gemma~2 9B    & 0.508 & 0.495 & 0   & $0.79^{***}$ \\
\bottomrule
\end{tabular}
\label{tab:continuation}
\end{table}

Template retrieval is ruled out for all models (agreement $\approx 0.50$). GPT-4's significant bigram correlation indicates statistical continuation---the model reads prefix statistics and reproduces them.

\subsection{Training Data Experiments}

\subsubsection{Corpus Search}

We searched for H/T sequences of length $\geq 10$ using regex matching.

\begin{table}[t]
\centering
\caption{Corpus search results. Genuine coin-flip sequences are essentially absent from pretraining corpora.}
\small
\begin{tabular}{lrc}
\toprule
Corpus & Docs Searched & Sequences Found \\
\midrule
C4 & 500,000 & 0 \\
Wikipedia & 500,000 & 2 \\
stats.stackexchange.com & 115,000 & 13 \\
Dolma v1.6 & 1,000,000 & 2 \\
\bottomrule
\end{tabular}
\label{tab:corpus-search}
\end{table}

The 13 stats.SE sequences (161 total flips) are themselves human-biased (alt rate 0.583, closer to Rapoport than Bernoulli), but this is negligible against trillions of training tokens.

\subsubsection{Infini-gram Analysis}

We queried the infini-gram API for exact n-gram counts across six indexed corpora.

\begin{table}[t]
\centering
\caption{Corpus bigram statistics show \textit{repetition} bias after H---the opposite of model behavior.}
\small
\begin{tabular}{lcccc}
\toprule
Corpus & $P(H|H)$ & $P(T|H)$ & $P(H|T)$ & $P(T|T)$ \\
\midrule
Dolma v1.7 & 0.648 & 0.352 & 0.501 & 0.499 \\
OLMo-mix & 0.743 & 0.257 & 0.498 & 0.502 \\
DCLM & 0.786 & 0.214 & 0.514 & 0.486 \\
RedPajama & 0.708 & 0.292 & 0.590 & 0.410 \\
Pile & 0.808 & 0.193 & 0.429 & 0.571 \\
C4 & 0.622 & 0.379 & 0.570 & 0.430 \\
\bottomrule
\end{tabular}
\label{tab:corpus-bigrams}
\end{table}

\textbf{Document-level analysis:} We retrieved 2,202 documents containing H/T 4-gram patterns from Dolma. Source distribution was $\sim$65\% code (GitHub, Jupyter), $\sim$31\% web/academic, $\sim$3\% StackExchange. Zero documents contained coin-flip keywords near the H/T sequences---all were code artifacts.

\subsubsection{HTHH Template Source}

We examined 4-gram prefix distributions in the 54 short sequences ($\geq 4$ flips) found in stats.SE:

\begin{table}[t]
\centering
\caption{4-gram prefix distribution in corpus sequences. HTHH is \textit{not} overrepresented.}
\small
\begin{tabular}{lcc}
\toprule
Prefix & Count & Fraction (Expected: 6.25\%) \\
\midrule
HHHH & 14 & 26\% \\
TTTT & 9 & 17\% \\
HHHT & 6 & 11\% \\
HHTT & 6 & 11\% \\
HTHH & 3 & 5.6\% \\
HTHT & 1 & 1.9\% \\
\bottomrule
\end{tabular}
\label{tab:prefix-distribution}
\end{table}

Run-starting prefixes (HHHH, TTTT) dominate; HTHH appears below the fair rate. Infini-gram counts confirm HTHH is consistently less common than HTHT across all corpora (ratio 0.32--0.65). The source of Llama 3's HTHH fingerprint remains unresolved.

\subsubsection{Clinical Text Analysis}

To investigate the \textsc{treatment} framing effect, we searched 115K stats.SE documents for clinical language and extracted binary sequences.

\begin{itemize}[nosep,leftmargin=*]
\item 16,611 clinical documents found (14.4\% of corpus)
\item 113 documents (0.7\%) contain explicit alternation language
\item 1,679 binary sequences extracted
\item Mean alternation rate: \textbf{0.326}---well below Bernoulli (0.500)
\end{itemize}

Clinical binary sequences are \textit{anti-alternating}. The \textsc{treatment} framing effect likely arises from procedural-balance language (``block randomization,'' ``every other patient'') rather than sequence statistics.

\subsection{Framing Perturbation}

We tested eight prompt framings across six models.

\begin{table*}[t]
\centering
\caption{Framing perturbation results. \textsc{treatment} elicits the highest alternation for all models.}
\small
\begin{tabular}{lccc}
\toprule
Model & Baseline & Min (Framing) & Max (Framing) \\
\midrule
GPT-4o-mini & 0.642 & 0.633 (bernoulli) & 0.660 (treatment) \\
Gemma 2 9B & 0.704 & 0.692 (random\_walk) & 0.770 (treatment) \\
GPT-4 & 0.620 & 0.620 (baseline) & 0.724 (treatment) \\
GPT-3.5 & 0.545 & 0.545 (baseline) & 0.666 (treatment) \\
Llama 3 8B & 0.573 & 0.564 (bitstring) & 0.793 (treatment) \\
Llama 3.1 8B & 0.584 & 0.576 (bitstring) & 0.801 (treatment) \\
& & & \\
\bottomrule
\end{tabular}
\label{tab:framing-perturbation}
\end{table*}

\textbf{Key findings:}
\begin{itemize}[nosep,leftmargin=*]
\item Above-chance alternation persists across \textit{all} framings and models.
\item \textsc{treatment} consistently produces the highest alternation.
\item Model-level variation in framing sensitivity is substantial (spread ranges from 0.027 to 0.315).
\end{itemize}
\end{document}